\documentclass[runningheads]{llncs}
\usepackage{graphicx}
\usepackage{amsmath,amssymb} 
\usepackage{color}
\usepackage[a-1b]{pdfx}
\usepackage[width=122mm,left=12mm,paperwidth=146mm,height=193mm,top=12mm,paperheight=217mm]{geometry}

\DeclareMathOperator*{\argmax}{arg\,max}

\begin{document}
\title{Visual Text Correction} 

\titlerunning{Visual Text Correction}
%
\author{Amir Mazaheri \and
Mubarak Shah}
%
\authorrunning{A. Mazaheri and M. Shah}
%

\institute{Center for Research in Computer Vision, University of Central Florida \\  \email{amirmazaheri@knights.ucf.edu ~~~~ shah@crcv.ucf.edu}}
\maketitle              
%


\begin{abstract}


This paper introduces a new problem, called {\em Visual Text Correction (VTC)}, i.e., finding and replacing an inaccurate word in the textual description of a video. We propose a deep network that can simultaneously detect an inaccuracy in a sentence, and fix it by replacing the inaccurate word(s).  Our method leverages the semantic interdependence of videos and words, as well as the short-term and long-term relations of the words in a sentence.  Our  proposed formulation can solve the VTC problem employing an End-to-End network in two steps: (1)Inaccuracy detection, and (2)correct word prediction. In detection step,  each word of a sentence is reconstructed such that the  reconstruction for the inaccurate word is maximized.
We exploit  both Short Term and Long Term Dependencies employing respectively  Convolutional N-Grams  and  LSTMs to reconstruct the word vectors. For  the  correction  step,  the  basic  idea  is  to  simply  substitute  the  word  with  the maximum reconstruction error for a better one. The second step is essentially a  classification  problem  where  the  classes  are  the  words  in  the  dictionary  as replacement options.
Furthermore, to train and evaluate our model, we propose an approach to automatically construct a large dataset for the VTC problem. Our experiments and performance analysis demonstrates that the proposed method provides very good results and also highlights the general challenges in solving the VTC problem. To the best of our knowledge, this work is the first of its kind for the Visual Text Correction task.

\end{abstract}

\section{Introduction}

Text Correction (TC) has been a major application of Natural Language Processing (NLP). Text Correction can be in form of a single word auto-correction system, which notifies the user of misspelled words and suggests the most similar word, or an intelligent system that recommends the next word of an inchoate sentence. In this paper, we  formulate a new type of Text Correction problem named {\em Visual Text Correction} ({\bf VTC}). In VTC, given a video and an inaccurate textual description in terms of a sentence about the video, the task is to fix the inaccuracy of the sentence. 

\begin{figure}
\begin{center}
   \includegraphics[width=1 \linewidth]{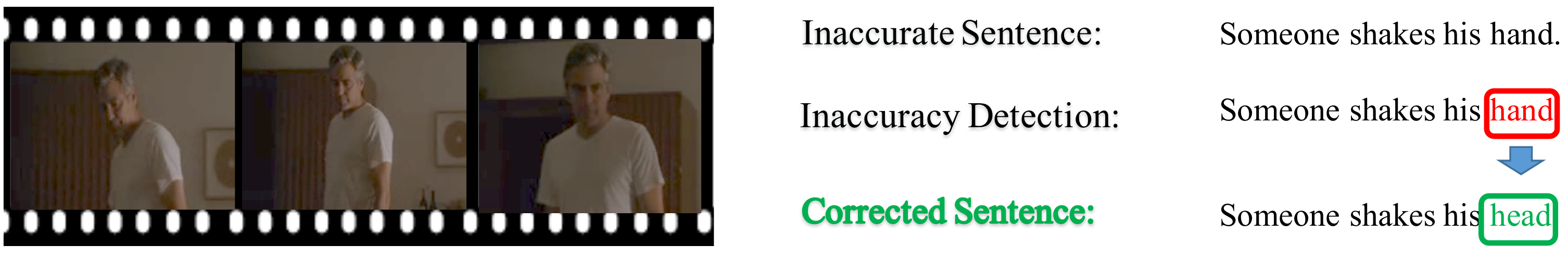}
\end{center}
   \caption{One inaccurate sentence example for a given video. The VTC task is to find the inaccuracy and replace it with a correct word.}
\label{fig:Concept}
\end{figure}

The inaccuracy can be in form of a phrase or a single word, and it may cause grammatical errors, or an  inconsistency in context of the given video. 
For example, the word ``car'' in the sentence: ``He is swimming in a car'' is causing a textual inconsistency and the word ``{\em hand}'' 
is causing an inaccuracy in the context of the video (See Figure ~\ref{fig:Concept}).

To formalize the problem, let sentence $\mathcal{S} = [w_1,w_2,...,w_N]$ consisting of $N$ words be an accurate description of the video $\mathcal{V}$, where $w_i \in \{0,1\}^{|V|}$, and $|V|$ is the number of words in our dictionary. For an inaccurate sentence $\widetilde{\mathcal{S}} = [\widetilde{w}_{1},\widetilde{w}_2,...,\widetilde{w}_N]$, the VTC task is to find the inaccurate word $\widetilde{w}_{t^*}$ where $1 \leq t^* \leq N$ and also to estimate the replacement word $w_{t}$. There can be several inaccurate words in a sentence; However, we train our system using sentences with just one inaccurate word. Nonetheless, we show that our trained network can be applied to sentences with multiple inaccurate words.

\begin{figure*}[ht]
\begin{center}
   \includegraphics[width=1 \linewidth]{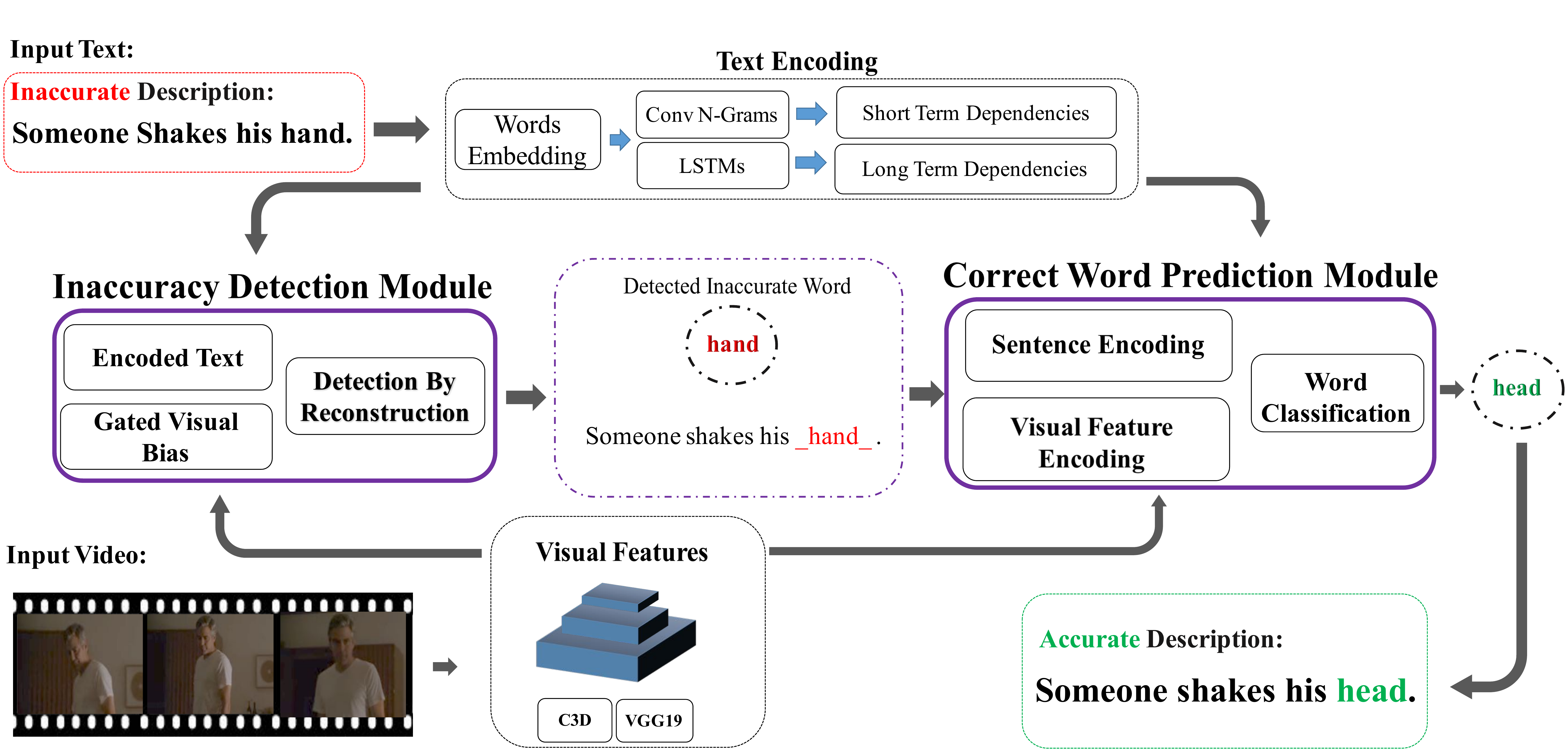}
\end{center}
   \caption{Proposed framework for  Visual Text Correction. The goal is to find and replace the inaccurate word in the descriptive sentence of a given video. 
   There are two main modules: 
   1) The Inaccuracy Detection module finds the inaccurate word, and 2) the Correct Word Prediction module predicts an accurate word as a substitution. Both of these modules use the encoded text and visual features. The Inaccuracy Detection uses  Visual Gating Bias to detect an inaccuracy and the Word Prediction Modules uses an efficient method to encode a sentence and visual features to predict the correct word.}
\label{fig:workFlow}
\end{figure*}

 Our  proposed formulation can solve the VTC problem employing an End-to-End network in two steps: (1)Inaccuracy detection, and (2)correct word prediction. Figure~\ref{fig:workFlow} shows the proposed framework of our approach. During the first step, we detect the inaccuracy by reconstruction, that is,  we embed each  word into a continuous vector, and  reconstruct a word vector for each of the words in the sentence based on its neighboring words. 
 A large distance between the reconstructed vector and the actual word vector implies an inaccurate word. For the second step, the basic idea is to simply substitute the word with the maximum reconstruction error for a better one. The second step is essentially a classification problem where the classes are the words in the dictionary as replacement options. 
\subsection{Motivations}
\textbf{Why Visual Text Correction?} We believe that the VTC  is very challenging and is a demanding problem to solve. During the last few years, the  integration of  computer vision and natural language processing (NLP)  has received a lot of attention, and excellent progress has been made. Problems like Video Caption Generation, Visual Question Answering, etc., are prominent examples of this progress. With this paper, we start a new  line of research which has many potential applications of VTC in real-world systems such as caption auto correction for video sharing applications and social networks, false tolerant text-based video retrieval systems, automatic police report validation,  etc.

\textbf{Why is VTC challenging?} Given a large number of words in a dictionary, many different combinations of words can take place in a sentence.  For example, there are ${|V|}\choose{3}$ possible triplet combinations of words from a dictionary of size $|V|$, which makes pre-selection of all possible correct combinations impractical. Also, in many cases, even a meaningful combination of words may result in an incorrect or inconsistent sentence. Furthermore, sentences can vary in length, and the inaccuracy can be in the beginning, middle or at the end of a sentence. Last but not least, a VTC approach must find the inaccuracy and also choose the best replacement to fix it. The video can provide useful information in addition to text since some words of the sentence, like verbs and nouns, need to be consistent with the video semantics like objects and actions present in the video.


\subsection{Contributions}
 The contribution of this paper is three-fold. First, we introduce the novel VTC problem. Second, we propose a principled approach to solve the VTC problem by decomposing the problem into inaccurate word detection and  correct word prediction steps. We propose a novel sentence encoder and a gating method to fuse the visual and textual inputs.  Third, we offer an efficient way to build a large dataset to train our deep network and conduct experiments. We also show that our method is applicable to sentences with multiple inaccuracies.
\section{Related Work}
In the past few years Deep Convolutional Neural Networks (CNNs) ~\cite{krizhevsky2012imagenet,Simonyan2014,szegedy2015going,he2015deep} have been demonstrated to be very useful in solving numerous Computer Vision problems like object detection ~\cite{deng2009imagenet,he2016deep}, action classification  ~\cite{soomro2012ucf101,tran2015learning}. Similarly, Recurrent Neural Networks (RNN)~\cite{schuster1997bidirectional,hochreiter1997long,chung2014empirical} and more specifically Long Short Term Memories(LSTM)~\cite{malinowski2015ask} have been influential in dramatic advances in solving many Natural Language Processing (NLP) problems such as Translation~\cite{gehring2017convolutional}, Paraphrasing~\cite{chen2011collecting}, Question Answering~\cite{bordes2015large,kumar2015ask,weston2014memory}, and etc. In addition to RNNs, several NLP works benefit from N-Grams~\cite{zhang2014kneser,chen1996empirical}, and convolutional N-Grams~\cite{kalchbrenner2014convolutional,gehring2017convolutional} to encode the neighborhood dependencies of the words in a sentence. The recent work  in ~\cite{gehring2017convolutional} show the superiority of N-Gram Convolutions over LSTM methods in sequence to sequence translation task. Therefore, in this paper we leverage N-Grams convolutions and Gating Linear Unit~\cite{dauphin2016language} in  encoding the text and also incorporating visual features in our inaccuracy detection network. In addition, studies on encoding semantics of words~\cite{mikolov2013efficient,Mikolov2013}, phrases and documents~\cite{le2014distributed,dai2015document} into vectors have been reported. The main goal of all these studies is to represent the textual data in a way that preserves the semantic relations. In this research, we use the representation and distance learning to reconstruct each word of a sentence and find the inaccurate word based on the reconstruction error.

NLP and CV advances have motivated a new generation of problems, which are at the intersection of NLP and CV. Image/Video captioning~\cite{vinyals2015show,johnson2015densecap,yu2017end} is to generate a description sentence about a given image/video. Visual Question Answering (VQA) ~\cite{Antol2015,Ren2015,malinowski2014multi,Ren2015,agrawal2015vqa,xiong2016dynamic,zhang2015yin} is to find the answer of a given question about a given image. In the captioning task, any correct sentence about the image/video can be acceptable, but in VQA, the question can be about specific details of the visual input. There are different types of the VQA problems, like multiple choice question answering~\cite{MovieQA}, Textbook Question Answering (TQA)~\cite{nadeem2017language}, Visual Dialogue~\cite{nadeem2017language}, Visual Verification~\cite{sadeghi2015viske}, Fill In the Blank (FIB)~\cite{maharaj2016dataset,yu2017end,mazaheri2017video}, etc. In addition to several types of questions in each of aforementioned works,  different kinds of inputs have been used. Authors in~\cite{MovieQA} introduced a dataset of movie clips  with the corresponding subtitles (conversations between actors) and questions about each clip. TQA~\cite{nadeem2017language} is a more recent form of VQA, where the input is a short section of elementary school textbooks including multiple paragraphs, figures, and a few questions about each. The aim of Visual Dialogue~\cite{nadeem2017language} is to keep a meaningful dialogue about a given photo, where a dialogue is a sequence of questions asked by a user followed by answers provided by system. Visual Knowledge Extraction~\cite{sadeghi2015viske} problem is to verify statements by a user (e.g. ``Do horses fly?'') from web crawled images.

Fill-In-the-Blank (FIB)~\cite{maharaj2016dataset,yu2017end,mazaheri2017video} is the most related to our work. FIB is a Question Answering task, where the question comes in the form of an incomplete sentence. In the FIB task, the position of the blank word in each sentence is given and the aim is to find the correct word to fill in the blank. Although FIB is somehow similar to the proposed VTC task, it is not straightforward to correct an inaccurate sentence with a  simple FIB approach. In FIB problem  the position of the blank is given, however in VTC it is necessary to find the inaccurate word in the sentence first and then substitute it with the correct word.   

Traditional TC tasks like grammatical and spelling correction have a rich literature in NLP. For instance, the authors in ~\cite{mays1991context} train a Bayesian network to find the correct misspelled word in a sentence. Other line of works like ~\cite{wu2010sentence,wagner1974order}, try to rephrase a sentence to fix a grammatical abnormality. In contrast to  works in ~\cite{mays1991context,suhm2001multimodal,wu2010sentence,wu2010sentence,wagner1974order},  there is no misspelled word in our problem, and we solve the VTC problem even for cases when the grammatical structure of the sentence is correct. Also, reordering the words of a sentence~\cite{wagner1974order} cannot be the solution to our problem, since we need to detect and replace a single word while preserving the structure of the sentence. Moreover, this is the first work to employ the videos in the Textual Correction task.


\section{Approach}
To formulate the VTC problem, assume $\widetilde{\mathcal{S}} = [\widetilde{w}_{1},\widetilde{w}_2,...,\widetilde{w}_N]$ is a given sentence for the video $\mathcal{V}$. Our aim is to find the index of the incorrect word, ${t^*}$,  and correct it with ${w^*}_{{t^*}}$ as follows:

\begin{equation} \label{eq:theory}
    (t^*, w^{*}_{t^*}) = \argmax_{{1 \leq t \leq N}, w_t \in \beta}{p((t,w_{t}) | \widetilde{\mathcal{S}}, \mathcal{V})},
\end{equation}
where $w_{i} \in \{ 0, 1\}^{|V|}$ is an one-hot vector representing the $i'th$ word of the sentence, $|V|$ is the size of our dictionary and $N$ is the length of the sentence. Also, $\beta \subseteq V$ represents the set of all potential substitution words. Since ${t^*}$ and ${w^*}_{{t^*}}$ are sequentially dependent, we decompose the Equation~\ref{eq:theory} into two sub-tasks:  Inaccurate word detection as:

\begin{equation} \label{eq:detectionTheory}
    {t^*} = \argmax_{1 \leq t \leq N}{p(t | \widetilde{\mathcal{S}}, \mathcal{V})},
\end{equation}
and the accurate word ${w^*}_{{t^*}}$ prediction as:
\begin{equation} \label{eq:FIBTheory}
    {w^*}_{{t^*}} = \argmax_{w \in \beta}{p(w | \widetilde{\mathcal{S}}, \mathcal{V}, {t^*} ~)}.
\end{equation}


\subsection{Inaccuracy Detection} \label{seq:InaccuracyDetection}
We propose detection by reconstruction method to find the most inaccurate word in a sentence, leveraging the semantic relationship between the words in a sentence. In our approach,  each word of a sentence is reconstructed such that the  reconstruction for the inaccurate word is maximized. For this purpose, we build embedded word vector $x_{i} \in \mathbb{R}^{d_x}$ for each corresponding word $w_i$ using a trainable lookup table $\theta_x \in \mathbb{R}^{ |V| \times d_x}$.
We exploit  both Short Term and Long Term Dependencies employing respectively  Convolutional N-Grams  and  LSTMs to reconstruct the word vectors.



\subsubsection{Short-Term Dependencies:} \label{seq:convnet}

Convolutional N-Gram networks\cite{gehring2017convolutional} capture the \emph{short-term} dependencies of each word surrounding. Sentences can vary in length, and a proper model should not be confused easily by long sentences. The main advantage of N-Gram approach is its robustness to disrupting words in long sentences, since it considers just a neighboring block around each word.

\begin{figure}
\begin{center}
   \includegraphics[width=1 \linewidth]{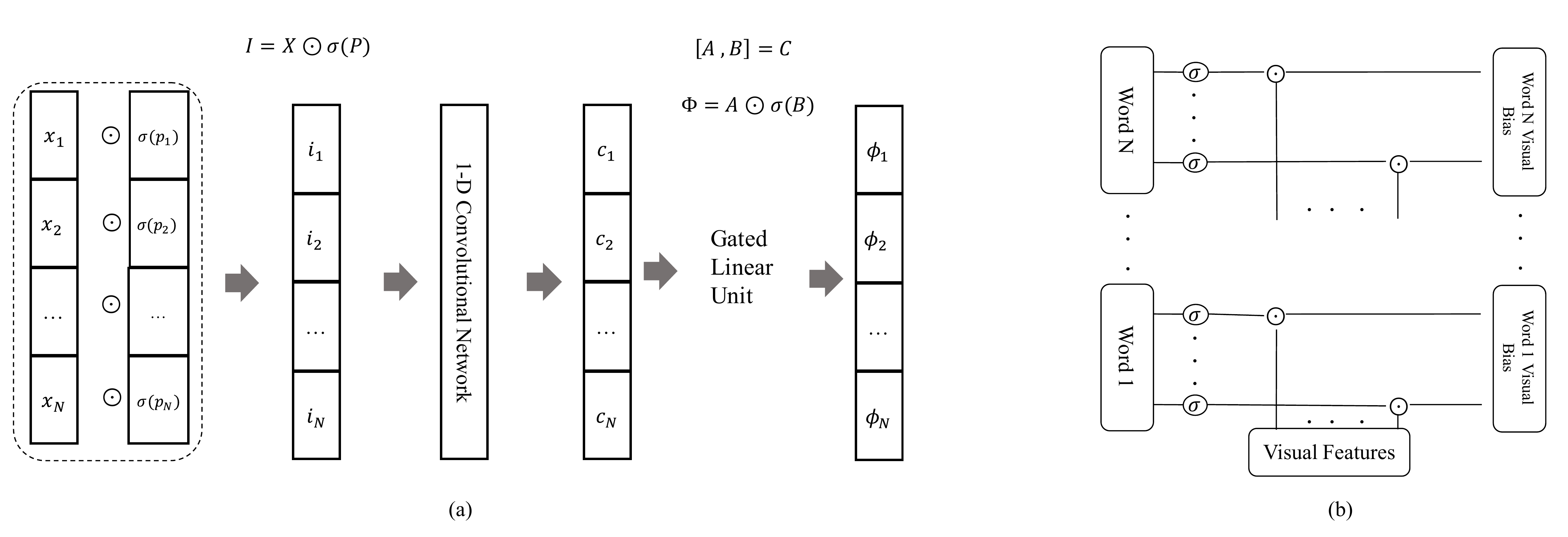}
\end{center}
   \caption{(a) One layer of Convolutional Text Encoding which captures the neighboring relationships. To extend one layer to multiple layers, we simply consider the $\phi_i$ vectors as $I_i$ for the next layer. (b) Our proposed Visual Gating Bias process. Given each word vector, we filter out some parts of a given visual feature through a gating process.}
\label{fig:ConvNGrams}
\end{figure}

Let $X = [x_{1}; x_{2}; \dots ;x_{N}]$ be the stacked vectors representing embedded word vectors.  Since the location of each word provides extra information about the correctness of that word in a sentence, we  combine it with word vectors $X$. We denote $p_t \in \mathbb{R}^{d_x}$ as an embedded vector associated to the $t$'th {\em position} of each sentence, which is one row of the trainable matrix,  $P \in \mathbb{R}^{N \times d_x}$.  We use $p_t$ values as gates for the corresponding word vectors $x_t$ for each sentence and get final combination $I$ as:
\begin{equation}\label{eq:PosGating}
I_t = x_t \odot \sigma (p_t),
\end{equation}
where $\odot$ denotes element-wise multiplication, and $I \in \mathbb{R}^{N \times d_x}$ is the input to a 1-D convolution with $2{d_x}$ filters and receptive field size of $m$. 
We call the resulting activation vectors $C \in \mathbb{R}^{N \times 2d_x}$. Furthermore, we use Gated Linear Units (GLU)~\cite{dauphin2016language} as the non-linear activation function. First, we split the $C$ matrix in half along its depth dimension:

\begin{equation} \label{eq:GLUDecompose}
\begin{split}
& [A, B] = C,\\
& \Phi = A \odot \sigma(B),
\end{split}
\end{equation}
where $A, B \in \mathbb{R}^{N \times d_x}$, and $\Phi = [\phi_{1}; \phi_{2};\dots;\phi_{N}]$, and $\phi_i \in \mathbb{R}^{d_x}$. The idea is to use the $B$ matrix as gates for the matrix A. An open gate lets the input pass, and a close gate changes the input to zero.
By stacking multiple 1-D convolutions and GLU activation functions the model goes deeper and the receptive field becomes larger. The output, $\Phi$, from each layer is the input, $I$, for the next layer. We call the final output $\Phi$, from the last Convolutional N-Grams layer, $\hat{X}^C \in \mathbb{R}^{N \times d_x}$. In Figure~\ref{fig:ConvNGrams}, we illustrate one layer of the N-Grams encoding.


\subsubsection{Long-Term Dependencies:} \label{seq:RNN}
Recurrent networks, and specifically LSTMs, have been successfully used to capture the \emph{long-term} relations in sequences. Long-term relations are beneficial to comprehend the meaning of a text and also to find the possible inaccuracies. To reconstruct a word vector based on the rest of the sentence using LSTMs, we define a left fragment and a right fragment for each word in a sentence. The left fragment starts from the first word of the sentence to one word before the word under consideration; and the right fragment is from the last word of the sentence to one word after the word under consideration in a reverse order. We encode each of the left and right fragments with a LSTM and extract the last hidden state vector of the LSTM as the encoded fragment:
\begin{equation} \label{eq:LSTMFormula}
    \hat{x}_{t}^{R} = W_{c} \times [u^{l}_t |  u^{r}_t],
\end{equation}
where $u^{l\text{/}r}_t \in \mathbb{R}^{h}$ are the encoded vectors of left/right fragments of the $t$'th word, and $W_{c} \in \mathbb{R}^{d_x \times 2h}$ is a trainable matrix to transform the $[u^{l}_t |  u^{r}_t]$ into the $\hat{x}^{R}_{t}$. 

\subsubsection{Detection Module:} \label{seq:DistanceModule}
We design a module to learn the distance between an actual word vector $x_t$ and the reconstructed $\hat{x}_t$ as explained above. 
This module learns to assign a larger distance to the inaccurate words and  reconstruct the predictions  as follows:

\begin{equation} \label{eq:distance}
    \mathcal{D}_t = W_d \times ( \dfrac{\hat{x}_t}{\| \hat{x}_t \|}  \odot \dfrac{{x}_t}{\| {x}_t \|}),
\end{equation}
 where $W_d \in \mathbb{R}^{1 \times d_x}$, and $\mathcal{D}_t$ is a scalar. $\hat{x}_t$ is the output of the text encoding; namely, $\hat{x}_{t} = \hat{x}^{C}_{t}$ for Convolutional N-Grams or $\hat{x}_{t} = \hat{x}^{R}_{t}$ in case of Recurrent Networks. Next, we combine both as a vector $\hat{x}_{t} = \hat{x}^{R}_{t} + \hat{x}^{C}_{t}$ to capture both long term and short term dependencies of a sentence. We design our distance module as a single layer network for simplicity; however, it can be a deeper network.

\subsubsection{Visual Features as Gated Bias:}


Visual features can contribute in finding the inaccuracy in a video description; however, it can be very challenging since some words may not correspond to any  visible form or shape (e.g. `weather'), while some others may correspond to distinct visual appearances (e.g. `cat'). 
We introduce a gating model to incorporate the visual features to measure the inconsistency of each word. The main idea is to find a dynamic vector for the visual features which changes for each word as follows (see Figure~\ref{fig:ConvNGrams}):
\begin{equation} \label{eq:visualNorm}
\Psi_\mathcal{V} = W_v \times \Omega(\mathcal{V}),
\end{equation}
where $\Omega(\mathcal{V}) \in \mathcal{R}^{d_v}$ is the visual feature  vector, and $W_v \in \mathcal{R}^{d_x \times d_v}$ is a transformation matrix for the visual features. We build the visual bias $v_t$ for each word vector $x_t$:
\begin{equation} \label{eq:visualgates}
v_t = \dfrac{\Psi_\mathcal{V}}{\| \Psi_\mathcal{V} \|} \odot \sigma( [W_g \times x_t]),
\end{equation}
 and $W_g \in \mathcal{R}^{d_x \times d_x}$ is transformation matrix, and $\| . \|$ denotes L2-Norm of a vector. The Sigmoid ($\sigma(.)$) operator bounds its input into $(0,1)$. It makes the model capable of refusing or accepting visual features dynamically for each word in a sentence.

The most intuitive way to incorporate the $V$ vectors in Equation~\ref{eq:distance}, is to use them as a bias term. In fact, the features which are refused by the word gates will have zero value and will act as neutral. Therefore, we use the following  updated form of Equation~\ref{eq:distance} with the video contribution:

\begin{equation} \label{eq:distanceplusvisual}
    \mathcal{D}_t = W_d \times ( \dfrac{\hat{x}_t}{\| \hat{x}_t \|}  \odot \dfrac{{x}_t}{\| {x}_t \|} \oplus v_t),
\end{equation}
where $\oplus$ denotes element-wise summation.
 
For the last step of the detection process, we find the word with maximum $\mathcal{D}$ value:
 
 \begin{equation}\label{eq:inaccuracySelection}
 t^* = \argmax_{ 1 \leq t \leq N}{(\mathcal{D}_t)}.
 \end{equation}

\subsubsection{Detection loss:}
We use the cross-entropy as detection loss function. Given the ground-truth one-hot vector $y \in \{ 0,1 \}^N$, which indicates the inaccurate word, and the $T^* = softmax(D)$ as probabilities, we compute the detection loss  $l_d$.


\subsection{Correct Word Prediction}
The second stage of our proposed method to solve the VTC problem is to predict a substitute word for the inaccurate word. Proposed correct word prediction consists of  three sub-modules: 1- Text Encoder, 2- Video Encoder, and 3- Inference sub-modules. 

\subsubsection{Text Encoder:}
This sub-module must encode the input sentence in such a way that the network be able to predict the correct word for the $t^*$'th word. We leverage the reconstructed word vectors $\hat{x}_t$ in equation~\ref{eq:distance}, since these vectors are rich enough to detect an inaccuracy by reconstruction error. We can feed the output of inaccuracy detection,  $t^*$, to our accurate word prediction network; however, the $argmax$ operator in Equation~\ref{eq:inaccuracySelection} is not differentiable and  prevents us to train our model End-to-End. To resolve this issue, we approximate the Equation~\ref{eq:inaccuracySelection} by vector $T^* = Softmax(D)$, which consists of probabilities of each of  $N$ words being incorrect in the sentence. We build the encoded text vector $q_t$:
\begin{equation} \label{eq:LRRLLSTMs}
q_t = tanh(W_q \times \hat{x}_t),
\end{equation}
where $W_q \in \mathbb{R}^{d_q \times d_x}$ is trainable matrix. $q_t \in \mathbb{R}^{d_q}$ is in fact a hypothetical representation of the textual description. To be more specific, $q_t$ is the encoded sentence, assuming that the word $t$ is the incorrect word, which is to be replaced by a blank, according to the Equation~\ref{eq:LRRLLSTMs}. Finally, the  \textbf{textual representation} $u_q \in \mathbb{R}^{d_q}$,  is  formulated as a weighted sum over all $q_t$ vectors:
\begin{equation}\label{eq:finalTextual}
u_q = \sum\limits_{t=1}^{N} {T^{*}_t}{q_t}.
\end{equation}

Note that, due to the ``$tanh(.)$'' operator in Equation~\ref{eq:LRRLLSTMs}, both $q_t$ and $u_q$ vectors have bounded values.

\subsubsection{Video Encoding:}
We leverage the video information to find the accurate word for $t^*$'th word of a sentence. While the textual information can solely predict a word for each location, visual features can help it to predict a better word based on the video, since the correct word can have a specific visual appearance. We extract the visual feature vector $\Omega(\mathcal{V})$ and compute our video encoding  using a fully-connected layer:
\begin{equation} \label{eq:spatialAttnWeights}
     u_{\mathcal{V}} = tanh({W}_{\mathcal{V}} \times \Omega (\mathcal{V})),
\end{equation}
where $W_{\mathcal{V}} \in \mathbb{R}^{d_q \times d_{v}}$, and $u_{\mathcal{V}} \in \mathbb{R}^{d_q}$ is our visual representation, which has bounded values. For simplicity, we have used just one layer video encoding; however, it can be a deeper and more complicated network.


\subsubsection{Inference:}
For the inference, we select the correct substitute word from the dictionary. In fact, this amounts to a classification problem, where the classes are the words and the inputs are the textual representation and the visual features:
\begin{equation} \label{eq:inference}
    w^{*}_{t^*} = \argmax_{w \in \beta} (W_i \times [u_q + u_{\mathcal{V}}]),
\end{equation}
where $W_i \in \mathbb{R}^{|\beta| \times d_q}$. 
Finally, we use cross-entropy to compute the correct word prediction loss, namely $l_f$. The total loss for our VTC method is $l = l_f + l_d$ and we train both sub-tasks together.

\section{Dataset and Experiments}
\subsection{Dataset} \label{seq:Dataset}
In this section, we describe our visual text correction dataset and the method to generate it. 
The main idea behind our approach to build a dataset for the VTC task is to remove one word from each sentence and substitute it with an inaccurate word; however, there are several challenges to address in order to build a realistic dataset. Here, we list a few  and also propose our approach to address those challenges. 

Our goal is to build a large dataset with a variety of videos with textual descriptions. We require that the vocabulary of the dataset and the number of video samples be large enough to train a deep network; hence we choose ``Large Scale Movie Description Challenge(LSMDC)'' dataset~\cite{maharaj2016dataset,rohrbach15cvpr}, which is one of the largest video description datasets available. Also, LSMDC has been annotated for ``Video Fill In the Blank (FIB)'' task. In FIB dataset, each video description contains one or more blanks, which needs to be filled in. For the VTC problem, we introduce inaccurate word in place of the blanks in FIB dataset. If there is more than one blanks in a sentence of the FIB dataset, we generate multiple examples of that sentence.


 Note that there are some important points related to selection of the replacement words, which we need to keep in mind. First, there shouldn't be a high correlation between the original and replacement words. For example, if we exchange the word ``car'' with ``bicycle'' frequently, any method will be biased and will always suggest replacing ``bicycle'' with ``car'' in all sentences. Second, we want our sentences to look natural even after the word substitution. Therefore, the replacement word should have the same ``Part Of Speech'' (POS) tag. For example, a singular verb is better to be replaced by another singular verb.

It is costly to manually annotate and select the replacement words for each sample, because of the significant number of videos, and the vast vocabulary of the dataset. Also, it is hard for the human annotators to prevent the correlation between the original and replacement words. We have considered all the mentioned points to build our dataset. Following we describe how we build a proper dataset for the VTC problem.

\subsubsection{Random Placement:}
In this approach, for each annotated blank in the LSMDC-FIB dataset, we place a randomly selected word from dictionary. This approach evidently is the most straightforward and simple way to introduce the incorrect word. However, in this method, a bias towards some specific words may exist, since the selected inaccurate words may not follow the natural distribution of the words in the dictionary. For example, we have many words with less than 4 or 5 occurrences in total. By Random Placement approach, rare words and the words with high frequencies have the same chance to show up as an inaccurate word.  This increases the rate of ``inaccurate occurrences to accurate occurrences'' for some specific words. This imbalanced dataset allows any method to detect the inaccuracy just based on the word itself not the the word in the context. Also, since replacement and original words may not take the same POS tag, Random Placement approach cannot meet one of the requirements mentioned above. 

\subsubsection{POS and Natural Distribution:}
 Due to the weaknesses of the Random Placement, we introduce a more sophisticated approach that selects the inaccurate words from a set of words with the same tag as the original (or accurate) word. 
 We first extract the POS tags of all the words from all the sentences using Natural Language Toolkit (NLTK)~\cite{loper2002nltk}, resulting in $32$ tags. Let $S_r$ be the set of all the words that takes the tag $r$ ($1 \leq r \leq 32$) at least once in the training sentences. To find a replacement for the annotated blank word $w$ with the tag $r$ in a sentence, we draw a sample from $S_r$ and use it as the inaccurate word. Obviously, some tags are more common than the others in natural language and as a result the incorrect words are similarly the same. 

To draw a sample from a set, we use the distribution of the words in all sentences. As a result, the words with more occurrences in the training set have more chance to be appeared as an inaccurate word. Therefore, the rate of incorrect to correct appearances of different words are close to each other. With this approach, we prevent the rare words to be chosen as the inaccurate word frequently and vice versa. 


\subsection{Results}
\subsubsection{Detection Experiments:}
In this subsection, we present our results for detection module and examine our method with various settings. The results are summarized  in Table~\ref{table:detectionResults}. Following we explain each experiment in more details.

\emph{Random} guess is to select one of the words in the sentence randomly as the inaccurate word. In \emph{Text Only Experiments} part of Table~\ref{table:detectionResults}, we compare all the blind experiments, where no visual features are used to detect the inaccuracy. \emph{Vanilla LSTM} uses a simple LSTM to directly produce the $\mathcal{D}_t$(Equation~\ref{eq:distance}) out of its hidden state using a fully connected layer.

\emph{One-Way Long-Term Dependencies} uses just $u_l$ in Equation~\ref{eq:LSTMFormula}. \emph{Long-Term Dependencies} experiment uses Recurrent Neural Networks method explained in Section~\ref{seq:RNN}. \emph{Convolutional N-Grams w/o Position Embedding} uses just Convolutional N-Grams, however, without the contribution of the positions of each word explained in Section~\ref{seq:convnet} while \emph{Convolutional N-Grams} is the complete explained module in Section~\ref{seq:convnet}. These two experiments show the effectiveness of our proposed words position gating, and finally, \emph{Convolutional N-Grams + Long-Term Dependencies} uses the combination of Convolutiona N-Grams and RNNs as mentioned in Section~\ref{seq:DistanceModule}. The last experiment reveals the contribution of both short-term and long-term dependencies of words in a sentence for the TC task. 

To further study  the strength of our method to detect the wrong words, we compare our method with a \emph{Commercial Web-App}\footnote{www.grammarly.com}. This application can detect structural or grammatical errors in text. We provide $600$ random samples from the test set to the web application and examine if it can detect the inaccuracy. In Table~\ref{table:detectionResults}, we show the comparison between our method and the aforementioned web application. This experiment shows the superiority of our results and also the quality of our generated dataset.

In \emph{Video and Text Experiments} part of the Table~\ref{table:detectionResults}, we show experiments with both video and text.  \textbf{Visual Gated Bias} experiment shows the capability of our proposed formulation to leverage the visual features in the detection sub-task. To show the superiority of our visual gating method, we conduct \emph{Visual Feature Concatenation} experiment. In this experiment, we combine the visual feature vector $\Omega(\mathcal{V})$ with each of the vectors ${x}_t$ and $\hat{x}_t$ in Equation~\ref{eq:distance} using concatenation and a fully connected layer. For these experiments, we have used the pre-trained C3D~\cite{tran2015learning} to compute the $\Omega(\mathcal{V})$. 


\begin{table}
\centering
\caption{Detection Experiments Results. For these experiments we just evaluate the ability of different models to localize the inaccurate word.}
\resizebox{\textwidth}{!}{
\begin{tabular}{|l|c|}
\hline
\textbf{Method}  & \textbf{Accuracy (\%)} \\
\hline
\hline
Random &  8.3 \\
\hline
\multicolumn{2}{l} {\textbf{Text Only Experiments}} \\
\hline
Commercial Web-App &
18.8   \\
\hline
Vanilla LSTM (One LSTM w/o Prop. Detection Formula)  & 28.0 \\
\hline
One-Way Long-Term Dependencies (One LSTM)  & 58.0 \\
\hline
Long-Term Dependencies (BiLSTM) & 67.2\\
\hline
Conv N-Grams w/o Position Embedding  & 66.8 \\
\hline
Conv N-Grams  & 69.0\\
\hline
Conv N-Grams + Long-Term Dependencies & \textit{72.5}\\
\hline
\multicolumn{2}{l} {\textbf{Video and Text Experiments}} \\
\hline
 Conv N-Grams + Long-Term Dependencies + Visual Feature Concatenation & 72.8\\
\hline
 Conv N-Grams + Long-Term Dependencies + Visual Gated Bias & \textbf{74.5}\\
\hline
\end{tabular}}

\label{table:detectionResults}
\end{table}





\subsection{Correction Experiments}
In Table~\ref{table:CorrectionResults}, we provide our results for the correction task. Note that, the correction task is composed of both inaccurate word detection and correct word predictions sub-tasks; thus, a correct answer for a given test sample must have the exact position of the inaccurate word and also the true word prediction ($(t^*$,$w^{*}_{t^{*}})$ in Equation~\ref{eq:theory}). 

\emph{Our Model - Just Text} experiment demonstrates our method performance with only  textual information. \emph{Our Model With C3D Features} uses both video and text, with C3D~\cite{tran2015learning} features as visual features. Similarly, \emph{Our Model With VGG19 Features} shows the results when  VGG19~\cite{Simonyan14c} features are the visual input. In \emph{Our Pre-trained Detection Model + Pre-Trained FIB}~\cite{mazaheri2017video} experiment we use our best detection model from Table~\ref{table:detectionResults} to detect an inaccurate word. We remove the inaccurate word and make an incomplete sentence with one blank. Then, we use one of the pre-trained state of the art FIB methods~\cite{mazaheri2017video}, which uses two staged Bi-LSTMs (LR/RL LSTMs) for text encoding + C3D and VGG19 features + temporal and spatial attentions, to find the missing word of the incomplete sentence. We show the superiority of our method which has been trained End-to-End. In both of detection (Table~\ref{table:CorrectionResults}) and  correction (Table~\ref{table:detectionResults}) tasks, there are accuracy improvements after including visual features.
We also report the Mean-Average-Precision (MAP) metric, to have a comprehensive comparison. To measure the MAP, we compute $N\times|\beta|$ scores for all the possible  $(t^*,{w^*}_{t^*})$.
\begin{table}
\centering
\caption{Text Correction Experiments Results. For the correction task, a model needs to successfully locate the inaccurate word and  provides the correct substitution.}
\resizebox{\textwidth}{!}{
\begin{tabular}{|l|c|c|}
\hline
\textbf{Method}  & \textbf{Accuracy (\%)} & \textbf{MAP (\%)}  \\
\hline
\hline
Random &  0.04 & $\simeq$0 \\
\hline
Vanilla LSTM - Just Text & 17.2 & 17.7 \\
\hline
Our Model - Just Text & 35.2 & 36.9\\
\hline
Our Pre-trained Detection Model + Pre-Trained FIB~\cite{mazaheri2017video} & 36.0 & 38.6\\
\hline
Our Model With C3D Features & 38.6 & 39.8 \\
\hline
Our Model With VGG19 Features & 38.8 & 40.1 \\
\hline
Our Model With VGG19 + C3D Features & \textbf{38.9} & \textbf{40.7} \\
\hline

\hline
\end{tabular}}
\label{table:CorrectionResults}
\end{table}

\subsection{Multiple Inaccuracies}
Here, we show that our method is capable of to be generalized to sentences with more than one inaccurate words. We conduct a new experiment with multiple inaccuracies in the test sentences and show the results in Table~\ref{table:detectionMultiResults}. In fact, we replace all the annotated blank words in the LSMDC-FIB test sentences with an inaccurate word. We assume that the number of inaccuracies, $k$, is given for each test sample, but the model needs to locate them. To select the inaccuracies in each sentence, we use the LSMDC-FIB dataset annotations. Note that in training we use sentences that contain just one inaccurate word, similar to previous experiments. During the test time, we modify the Equation~\ref{eq:inaccuracySelection} to $t^*_{i =1,..,k} = arg\, kmax(\mathcal{D}_t)$, where $arg\, kmax$ returns the top $k$ inaccurate word candidates. Number of inaccurate words in our test set sentences reaches up to $10$ words. However, in Table~\ref{table:detectionMultiResults}, we show the detection results for sentences with each $k \leq 4$ value separately, and also the overall accuracy for all the $k$ values.
\begin{table}[t]
\centering
\caption{Detection and Correction results for sentences with multiple inaccuracies. Two types of Accuracy evaluations are provided. (1) Word-Based (WB) Accuracy: All correctly fixed incorrect words are counted independently. (2) Sentence-Based (SB) Accuracy: All inaccurate words in a sentence must be fixed correctly. Similarly, two types of MAP is reported: (1) WB-MAP, in which, one AP per each incorrect word is computed. (2) SB-MAP, in which, one AP per each sentence, including all the $k$ incorrect words, is computed.  $k$ represents the number of inaccuracies in each sentence.}
\resizebox{\textwidth}{!}{
\begin{tabular}{ |c | c c c c |c || c c c c |c| }
\hline
 k = & 1 & 2 & 3 & 4 & All & 1 & 2 & 3 & 4 & All \\ 
\hline
\# Of Test Samples & 1805& 4856 & 5961& 520 & 30349 & 1805 & 2428 & 1987 & 130 & 9575\\
\hline
\hline
  Detection & \multicolumn{5}{c ||}{WB-Acc. (\%)} & \multicolumn{5}{c|}{SB-Acc. (\%)}\\
\hline
Vanilla LSTM - Just Text  &59 & 63  &67 & 68 & 66 & 59 & 37 & 27 & 18 & 36 \\
\hline
 Our Method - Just Text &80 & 81 &80 & 80 & 80 & 80 & 65 & 48 & 37 & 59 \\
\hline
{\bf Our Method - Text + Video} &\textbf{85} &\textbf{83} & \textbf{83} & \textbf{82} & \textbf{83} & \textbf{85} & \textbf{68} & \textbf{54} & \textbf{39} & \textbf{63} \\
\hline
 Correction & \multicolumn{5}{c ||}{WB-Acc. (\%)} & \multicolumn{5}{c|}{SB-Acc. (\%)}\\
\hline
Our Method - Just Text &19 & 12 &12 & 11 & 3 & 19 & 2 &  $\simeq 0$ &  $\simeq 0$ & 5 \\
\hline
{\bf Our Method - Text + Video} &\textbf{24} & \textbf{18} &\textbf{17} & \textbf{17} & \textbf{18} & \textbf{24} & \textbf{4} & $\simeq 0$ & $\simeq 0$ & \textbf{7} \\
\hline
 Correction & \multicolumn{5}{c ||}{WB-MAP (\%)} & \multicolumn{5}{c|}{SB-MAP (\%)}\\
\hline
Our Method - Just Text &30 & 14 &10 & \textbf{8} & 12 & 30 & 15 & 11 & 9 & 17 \\
\hline
{\bf Our Method - Text + Video} &\textbf{35} & \textbf{17} &\textbf{11} & 7 & \textbf{14} & \textbf{35} & \textbf{18} & \textbf{12} & \textbf{10} & \textbf{19} \\
\hline
\end{tabular}}
\label{table:detectionMultiResults}
\end{table}

\begin{figure*}
\begin{center}
  \includegraphics[width=0.95 \linewidth]{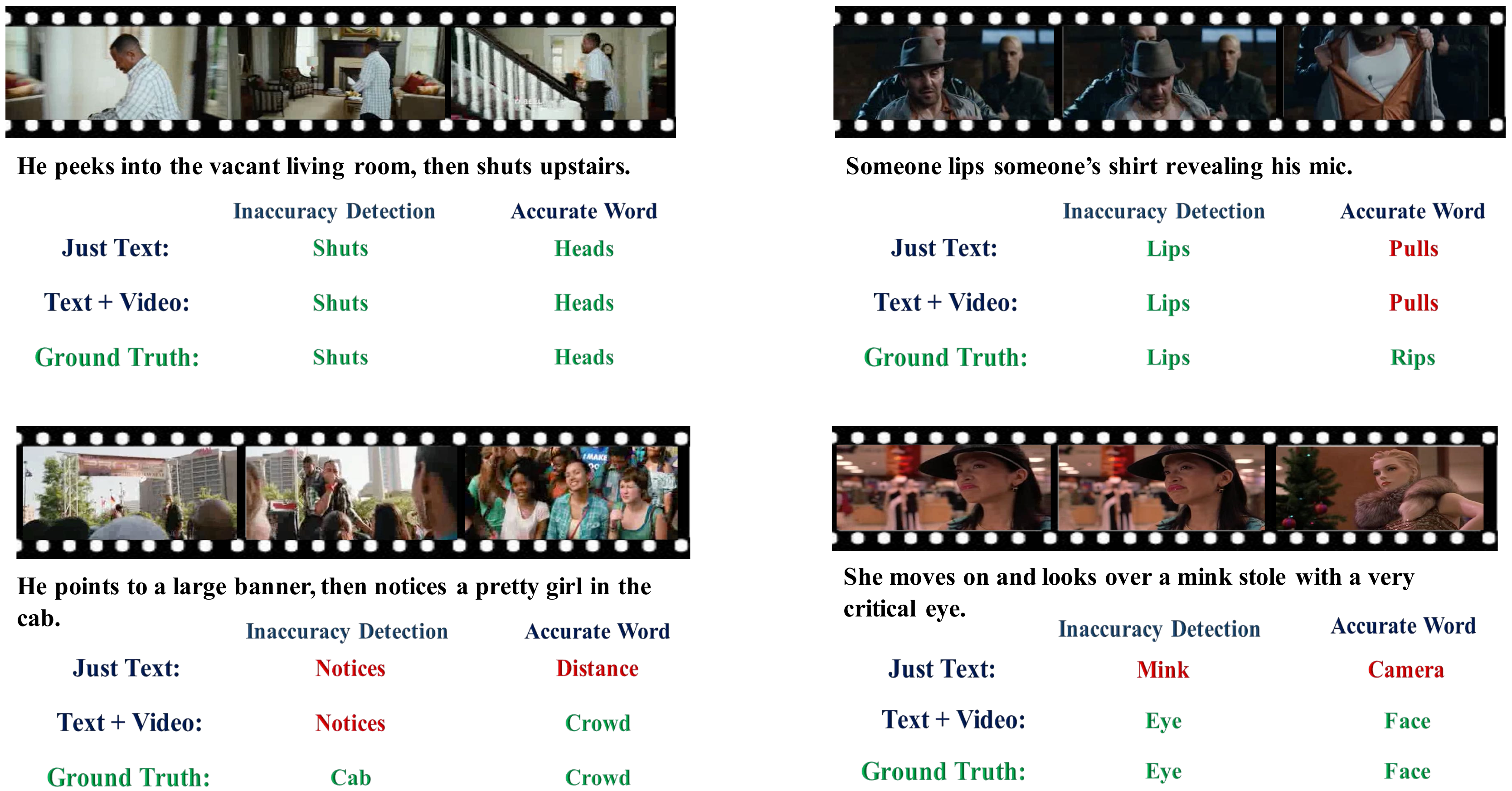}
\end{center}
  \caption{Here we show four samples of our test results. For each sample, we show a video and an inaccurate sentence,  the detected inaccurate word, and the predicted accurate word for substitution. The green color indicates a correct result while the red color shows a wrong result.}
\label{fig:qualitative}
\end{figure*}

\subsection{Qualitative Results}
We show a few VTC examples in Figure~\ref{fig:qualitative}. For each sample, we show frames of a video and corresponding sentence with an inaccuracy. We provide the qualitative results for each example using our ``Just Text'' and ``Text + Video'' methods. We show two columns for the detection and correct word prediction. The green and red colors respectively indicate true and false outputs. Note that, for the VTC task, just a good detection or prediction is not enough. Both of these sub-tasks are needed to solve the VTC problem. For example, the left bottom example in Figure~\ref{fig:qualitative} shows a failure case for both ``Just Text'', and ``Text + Video'', although the predicted word is correct using ``Text + Video''.

\section{Conclusion}
We have presented a new formulation of text correction problem, where the goal is to find an inaccuracy in a video description, and fix it by replacing the inaccurate word. We propose a novel approach to leverage  both  textual  and visual features to detect and fix the inaccurate sentences, and we show the superior results are obtained our approach. Moreover, we introduce an approach to generate a suitable dataset for VTC problem. Our proposed method provides a strong baseline for inaccuracy detection and correction tasks for sentences with one or multiple inaccuracies. We believe that our work is a  step forward in the research related to intersection of Natural Language Processing and Computer Vision. 
We hope that this work lead to more exciting future researches in VTC.

{\bf ACKNOWLEDGMENTS}
This material is based upon work supported by the National
Science Foundation under Grant No. 1741431. Any opinions,
findings, and conclusions or recommendations expressed in
this material are those of the author(s) and do not necessarily
reflect the views of the National Science Foundation.

\bibliographystyle{splncs}
\bibliography{egbib}
\end{document}